\documentclass[10pt,twocolumn,letterpaper]{article}

\usepackage{cvpr}
\usepackage{times}
\usepackage{epsfig}
\usepackage{graphicx}
\usepackage{amsmath}
\usepackage{amssymb}
\usepackage{enumitem}
\usepackage{boldline}
\usepackage{subfig}
\usepackage{multirow}
\usepackage[dvipsnames]{xcolor}
\usepackage{pifont}
\usepackage{placeins}
\usepackage{booktabs}
\usepackage{multicol}

\usepackage[pagebackref=true,breaklinks=true,letterpaper=true,colorlinks,bookmarks=false,citecolor=ForestGreen]{hyperref}
\usepackage[square,numbers,sort&compress]{natbib}
\usepackage[font=footnotesize,labelfont=bf]{caption}

\cvprfinalcopy 
\def\cvprPaperID{4} 

\ifcvprfinal\fi
\begin{document}

\title{Does Object Recognition Work for Everyone?}

\author{
Terrance DeVries\thanks{Equal contribution.}
\qquad
Ishan Misra\footnotemark[1]
\qquad
Changhan Wang\footnotemark[1]
\qquad
Laurens van der Maaten\\
Facebook AI Research\\
{\tt\small \{tldevries,imisra,changhan,lvdmaaten\}@fb.com}
}

\maketitle

\begin{abstract}
The paper analyzes the accuracy of publicly available object-recognition systems on a geographically diverse dataset. This dataset contains household items and was designed to have a more representative geographical coverage than commonly used image datasets in object recognition. We find that the systems perform relatively poorly on household items that commonly occur in countries with a low household income. Qualitative analyses suggest the drop in performance is primarily due to appearance differences within an object class (e.g., dish soap) and due to items appearing in a different context (e.g., toothbrushes appearing outside of bathrooms). The results of our study suggest that further work is needed to make object-recognition systems work equally well for people across different countries and income levels.
\end{abstract}

\section{Introduction}
\label{sec:introduction}

\begin{figure}
\centering
\includegraphics[width=0.95\linewidth]{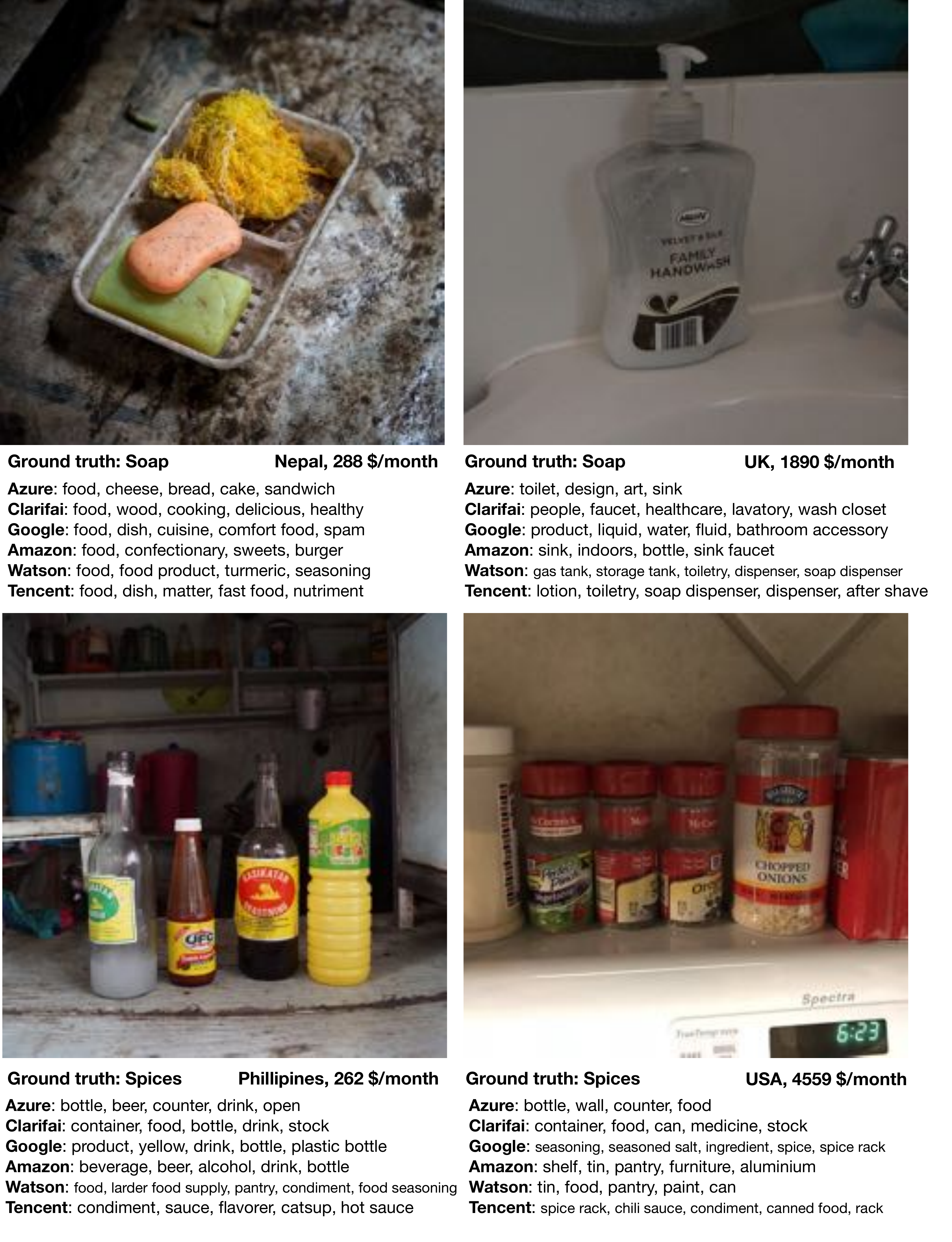}
\caption{Images of household items across the world, and classes recognized in these images by five publicly available image-recognition systems. Image-recognition systems tend to perform worse in non-Western countries and for households with lower incomes. See supplemental material for license information.}
\label{fig:teaser}
\end{figure}

Recent advances in the accuracy of object-recognition systems \cite{he2016,huang2017,szegedy2017} have spurred a large variety of real-world deployments of such systems, for instance, in aids for the visually impaired \cite{gurari2018}, in photo album organization software \cite{wang2017}, in image search, and in popular cloud services \cite{amazonrekognition,clarifai,googlecloudvision,ibmwatson,microsoftazure}. With the great success of such deployments also comes great responsibility: in particular, the responsibility to ensure that object-recognition systems work equally well for users around the world, irrespective of their cultural background or socio-economic status.

This paper investigates whether \emph{current object-recognition systems work well for people across countries and income levels}. Our study suggests these systems are less effective at recognizing household items that are common in non-Western countries or in low-income communities. When used to recognize such items, the error rate of object-recognition systems for households with an income of less than US\$$50$ per month is approximately $10\%$ lower compared to households making more than US\$$3,500$ per month; for some models, the difference is even larger. Similarly, the \emph{absolute} difference in accuracy of recognizing items in the United States compared to recognizing them in Somalia or Burkina Faso is around $15\!-\!20\%$. These findings are consistent across a range of commercial cloud services for image recognition. 

Figure~\ref{fig:teaser} shows two pairs of examples of household items across the world and the classifications made by five publicly available image-recognition systems. The results of our study suggest additional work is needed to achieve the desired goal of developing object-recognition systems that work for people across countries and income levels.

\section{Measuring Recognition Performance of Household Items Across the World}
\label{sec:measuring_errors}

\begin{figure}[t]
\centering
\includegraphics[width=\linewidth]{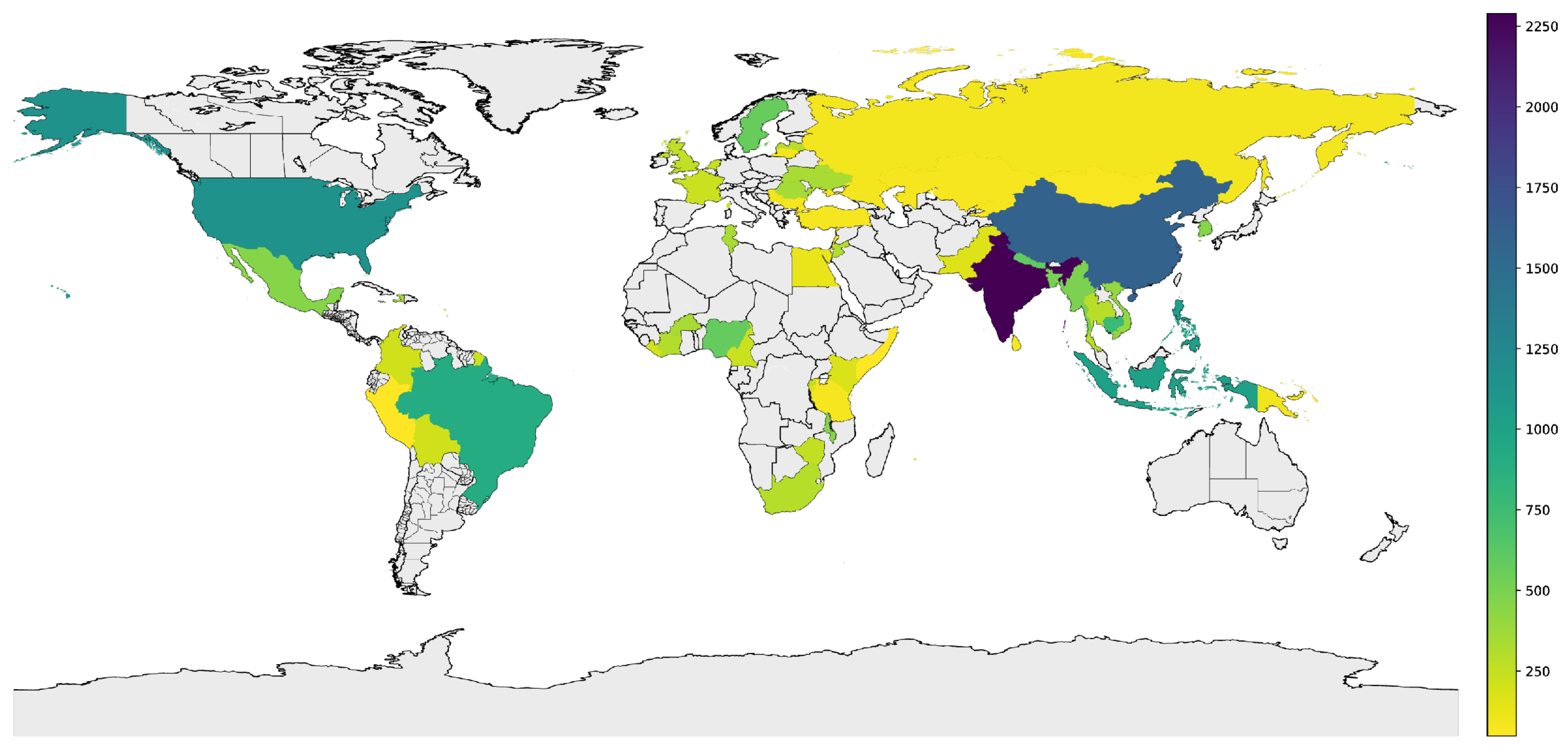}
\caption{Choropleth map displaying the number of images per country in the Dollar Street test set.}
\label{fig:Dollar Street_chloropleth}
\end{figure}

\noindent \textbf{Dataset.} We perform object-classification experiments on the Dollar Street image dataset of common household items. The Dollar Street dataset was collected by a team of photographers with the goal of making ``everyday life on different income levels understandable''~\cite{dollarstreet}. The dataset contains photos of $135$ different classes taken in $264$ homes across $54$ countries. Examples of images in the dataset are shown in Figure~\ref{fig:teaser}. The choropleth map in Figure~\ref{fig:Dollar Street_chloropleth} shows the geographic distribution of the photos in the dataset.

Some of the classes in the Dollar Street dataset are abstract (for instance, \emph{``most loved item''}); we remove those classes from the dataset and perform experiments on the remaining $117$ classes. The list of all classes for which we analyzed object-recognition accuracy is presented in Table~\ref{table:class_list}.

In addition to class annotations, the Dollar Street dataset contains information on: (1) the country in which the photograph was collected; and (2) the monthly consumption income of the photographed family in dollars adjusted for purchasing power parity (PPP). A detailed description on how this \emph{normalized monthly consumption income} was computed is presented in~\cite{dollarstreet}. We use both the location and the income metadata in our analysis.

\noindent \textbf{Experimental setup.} We measure the accuracy of five object-recognition systems that are publicly available through cloud services, namely, the systems provided by Microsoft Azure~\cite{microsoftazure}, Clarifai~\cite{clarifai}, Google Cloud Vision~\cite{googlecloudvision}, Amazon Rekognition~\cite{amazonrekognition}, and IBM Watson~\cite{ibmwatson}. We tested the versions of these systems that were publicly available in February 2019. In addition to the cloud-based systems, we also analyzed a state-of-the-art object recognition system that was trained exclusively on publicly available data: namely, a ResNet-101 model~\cite{he2016} that was trained on the Tencent ML Images dataset~\cite{wu2019} and achieves an ImageNet validation accuracy of $78.8\%$ (top-1 accuracy on a single $224 \times 224$ center crop).

\begin{figure}[h]
\centering
\includegraphics[width=\linewidth]{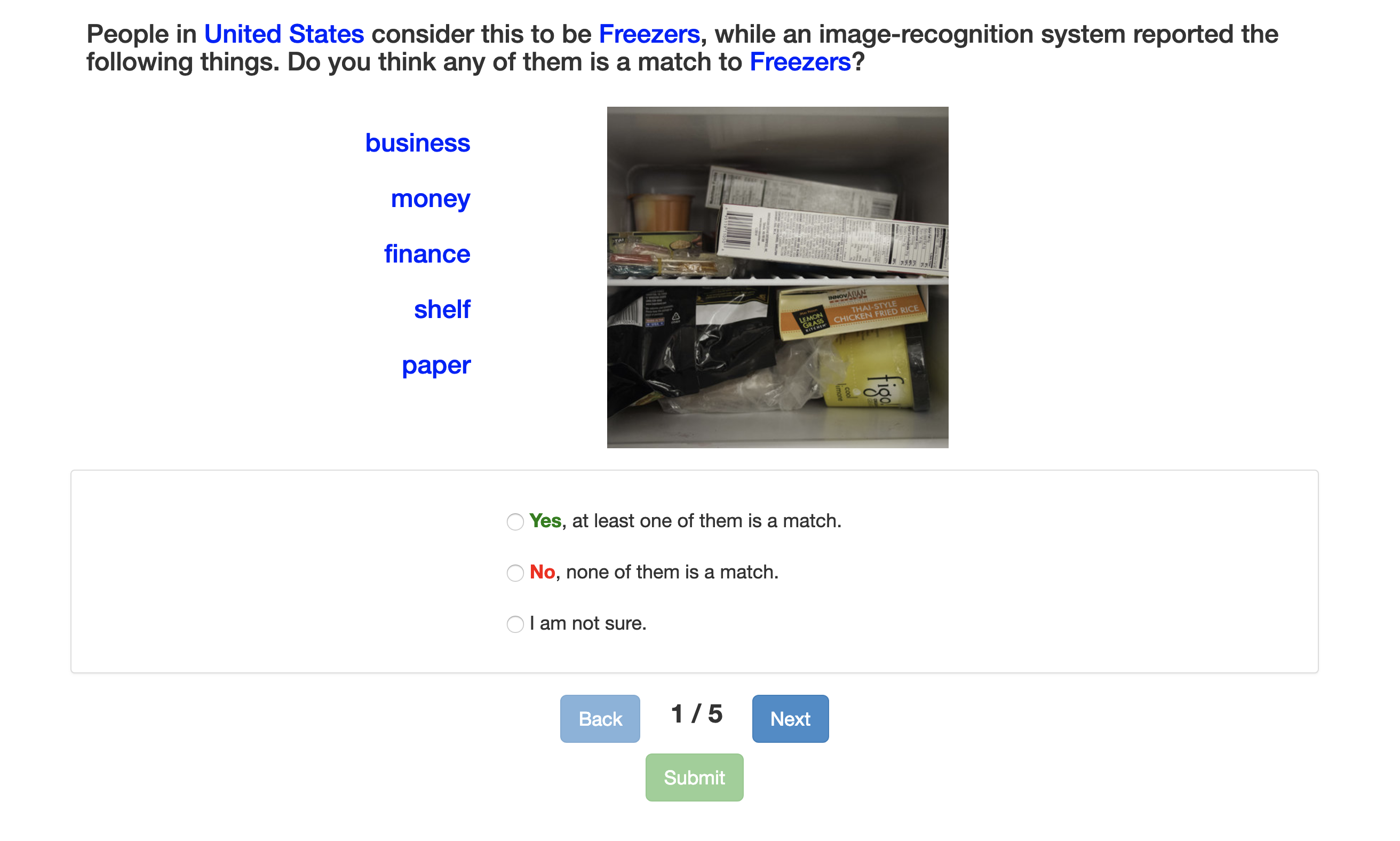}
\caption{Interface presented to human annotators tasked with assessing the correctness of predictions made by the image-recognition models.}
\label{fig:mturk_ui}
\end{figure}

We evaluate the quality of the predictions produced by all six systems in terms of accuracy@5 per assessment by human annotators\footnote{Due to external constraints, all annotators that participated in our study were based in the United States. Whilst this may bias the annotations, qualitative evaluations suggest that the impact of these biases is very small.}. Specifically, in order to determine if a prediction was correct, we asked human annotators to assess whether or not any of the five front-ranked predictions matched the ground-truth class annotation provided in the Dollar Street dataset. Figure~\ref{fig:mturk_ui} shows the annotation interface used; note that we did show the annotators the relevant photo as additional context on the annotation task. We report accuracies averaged over all six systems.

\begin{figure}[t]
\centering
\includegraphics[width=\linewidth]{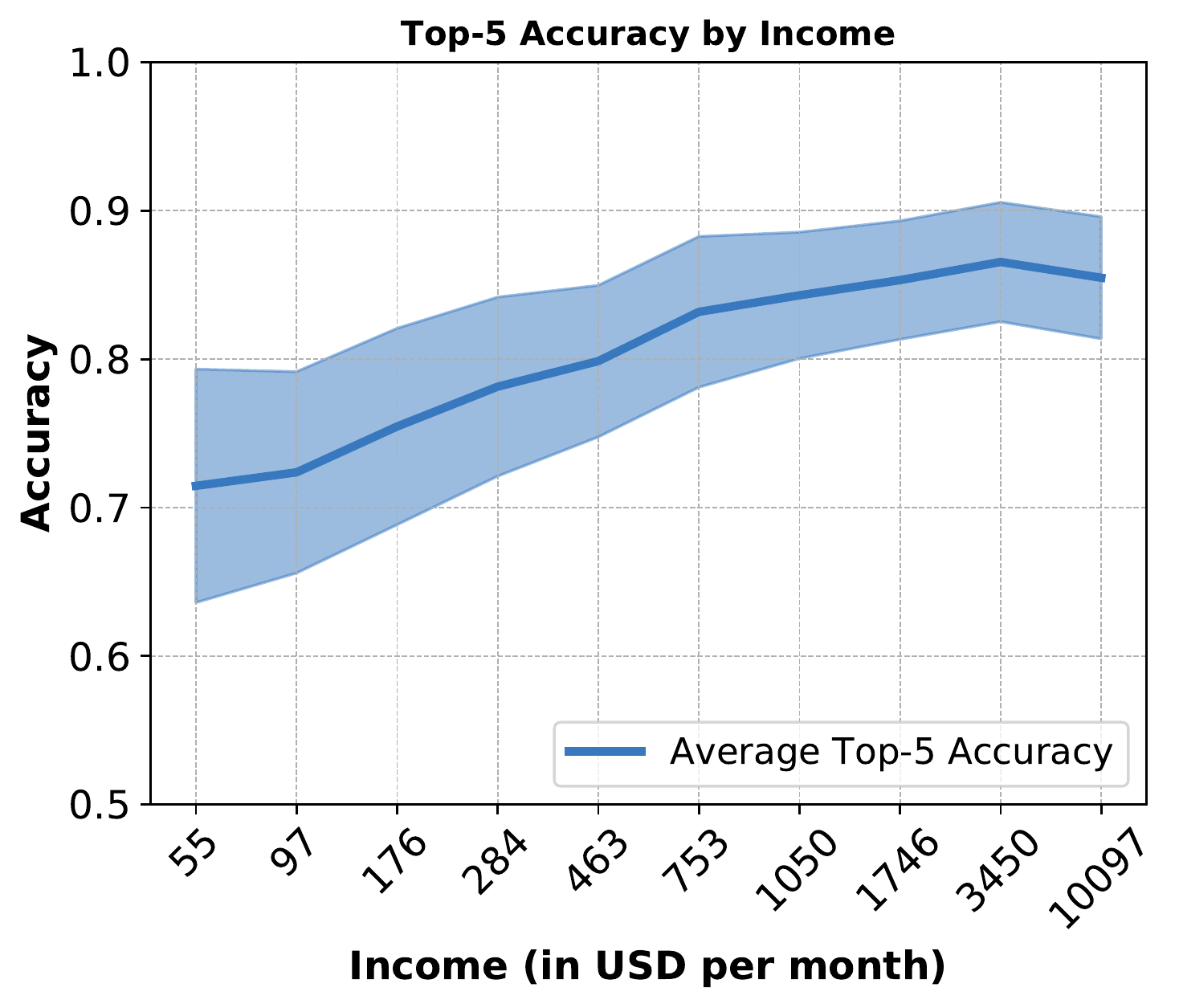}
\caption{Average accuracy (and standard deviation) of six object-recognition systems as a function of the normalized consumption income of the household in which the image was collected (in US\$ per month).}
\label{fig:income_plots_raw}
\vspace{-0.6em}
\end{figure}

\noindent \textbf{Results.} The average accuracy of the six object-classification systems on the Dollar Street dataset is shown in Figure~\ref{fig:income_plots_raw} as a function of the normalized monthly consumption income (PPP) of the household in which the photo was collected. We ensured that the number of images in each `income bin' are roughly the same ($2372 \pm 50$ images each) so that the accuracies per income bin are easily comparable. Whilst the exact accuracies vary somewhat per model, the results show the same pattern for all six systems: the object-classification accuracy in recognizing household items is substantially higher for high-income households than it is for low-income households. For all systems, the difference in accuracy for household items appearing the in the lowest income bracket (less than US\$$50$ per month) is approximately $10\%$ lower than that for household items appearing the in the highest income bracket (more than US\$$3,500$ per month). Figure~\ref{fig:teaser} sheds some light on the source of this discrepancy in accuracies: it suggest the discrepancy stems from household items being very different across countries and income levels (\emph{e.g.}, dish soap) and from household items appearing in different contexts (\emph{e.g.}, toothbrushes appearing in households without bathroom).

\begin{figure}[t]
\centering
\includegraphics[width=\linewidth]{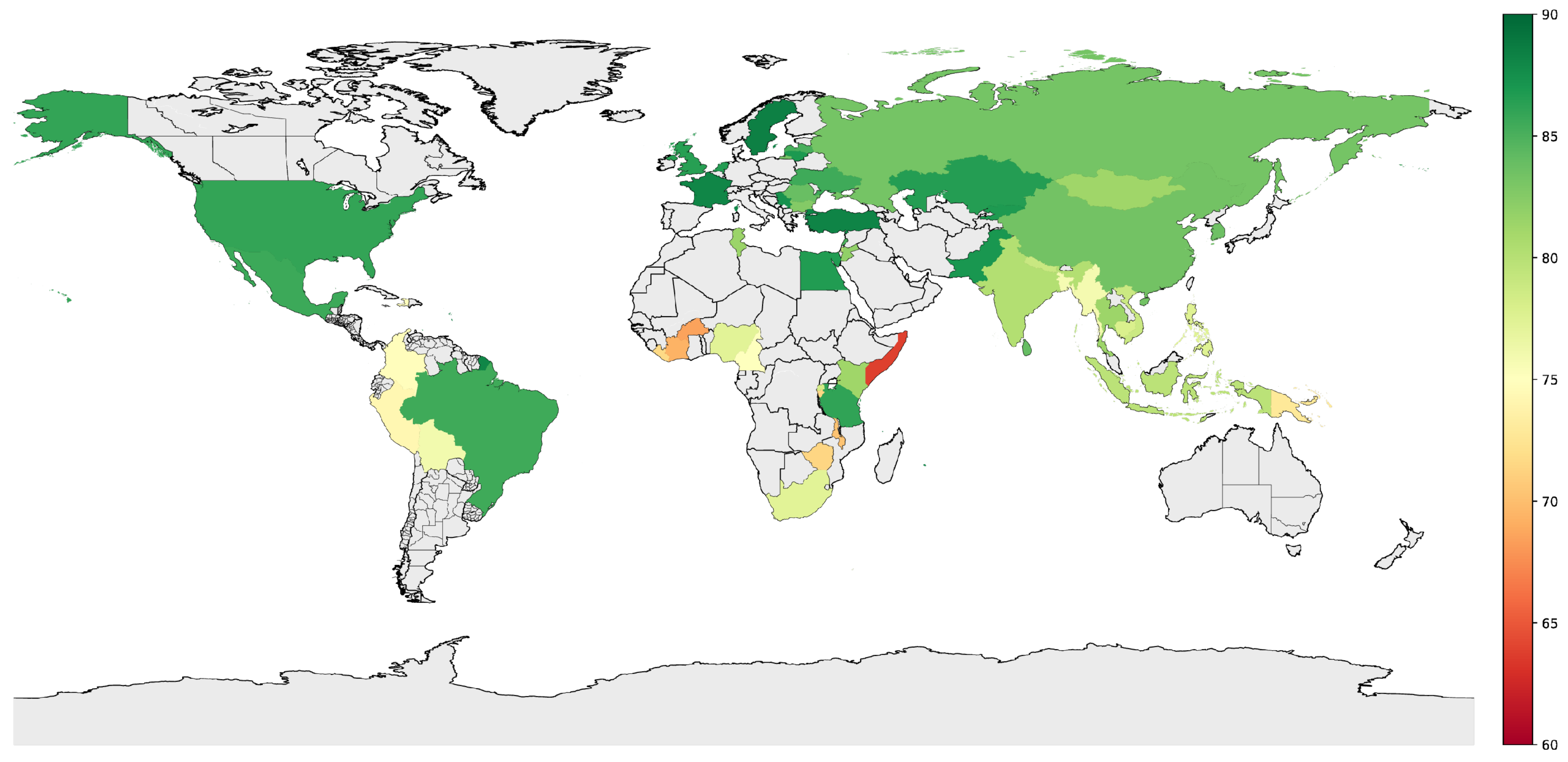}
\caption{Choropleth map displaying the average accuracy of six object-classification systems per country. The color red indicates an accuracy of $\sim\!60\%$, yellow an accuracy of $\sim\!75\%$, and green an accuracy of $\sim\!90\%$.}
\label{fig:rekog_chloropleth}
\end{figure}

Figure~\ref{fig:rekog_chloropleth} displays the average accuracy of the six object-classification systems as a function of geographical location in a choropleth map. The results highlight the differences in accuracies across countries. In particular, the accuracy of the systems is approximately $15\%$ (absolute) higher on household items photographed in the United States than it is on household items photographed in Somalia or Burkina Faso. We present additional results in the appendix.

\section{Sources of Accuracy Discrepancies}
\label{sec:analyzing_errors}

There are at least two causes for the observed discrepancies in object-classification accuracies: (1) the geographical sampling of image datasets is unrepresentative of the world population distribution and (2) most image datasets were gathered using English as the ``base language''. We also note that the Dollar Street dataset has a few classes where the images are labeled according to the affordance of the label, \eg, ``refrigerators'' for images from lower income groups may refer to pots and pans used to cool things.

\begin{enumerate}[leftmargin=*, noitemsep]

\item \textbf{Geographical distribution.} We analyze the geographical distribution of three popular computer-vision datasets: ImageNet~\cite{russakovsky2015}, COCO~\cite{coco}, and OpenImages~\cite{kuznetsova2018}. These datasets do not contain geographical information, but we can obtain the geographical information for the subset of the dataset images that originate from Flickr via the Flickr API. Figure~\ref{fig:density_maps} displays world maps with the resulting approximate geographical distribution of the ImageNet, COCO, and OpenImages datasets. For reference, the figure also displays a world population density map based on publicly available data from the European Commission JRC \& DGRD \cite{ecpopulation}. In line with prior analyses~\cite{shankar2017}, the density maps demonstrate that the computer-vision dataset severely undersample visual scenes in a range of geographical regions with large populations, in particular, in Africa, India, China, and South-East Asia. Whilst  income distribution correlates with geographical distribution, results in the supplementary material suggest income distribution is a driver for our results in itself, too (see Figure~\ref{fig:income_india}).

\begin{figure}[t]
\centering
\includegraphics[width=\linewidth]{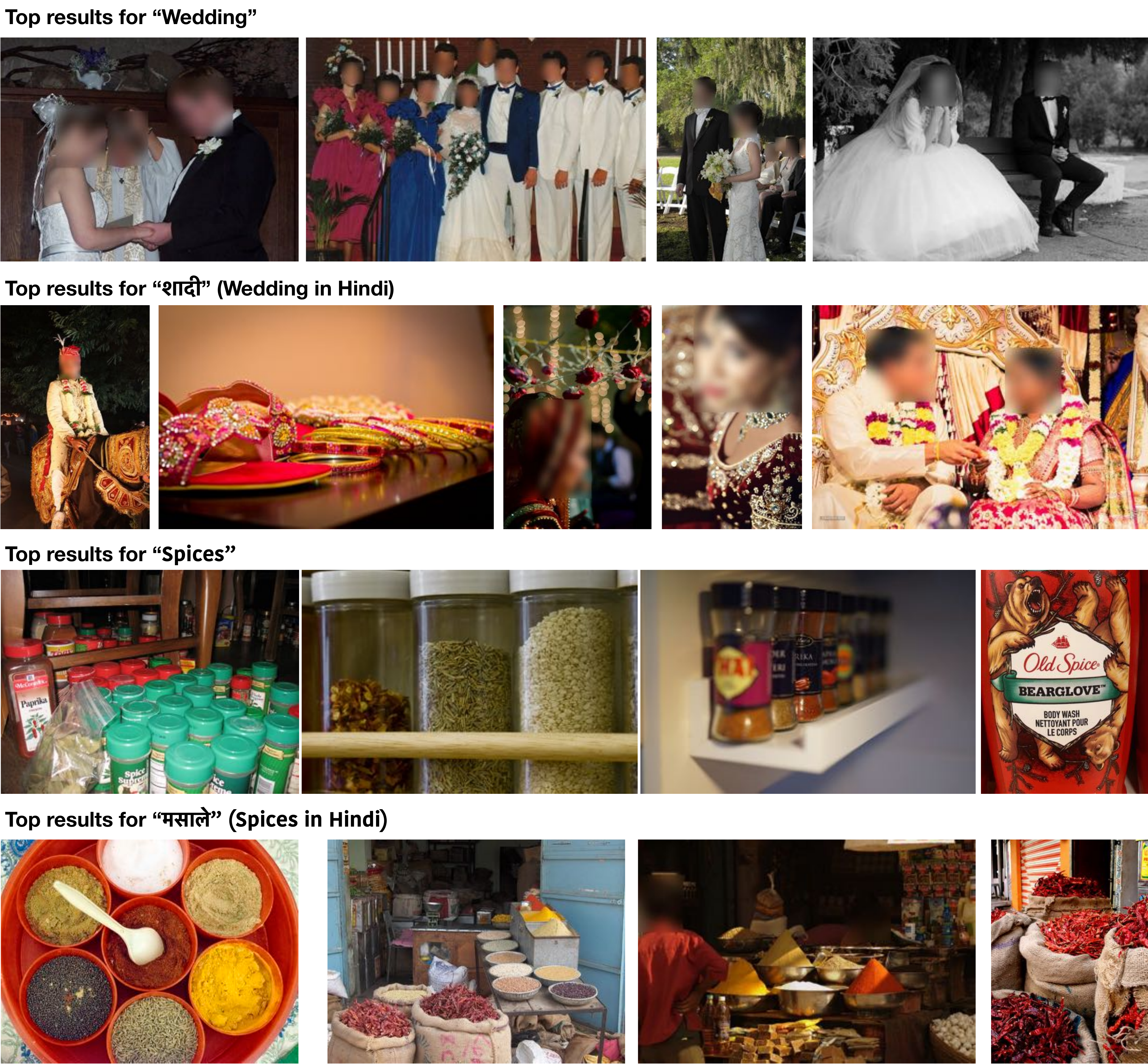}
\caption{Top Flickr results for the same queries in Hindi and English. The results returned across languages are visually different. See supplemental material for license information.}
\label{fig:flickr_queries}
\end{figure}

\begin{figure*}[t]
\centering
\includegraphics[width=0.95\textwidth]{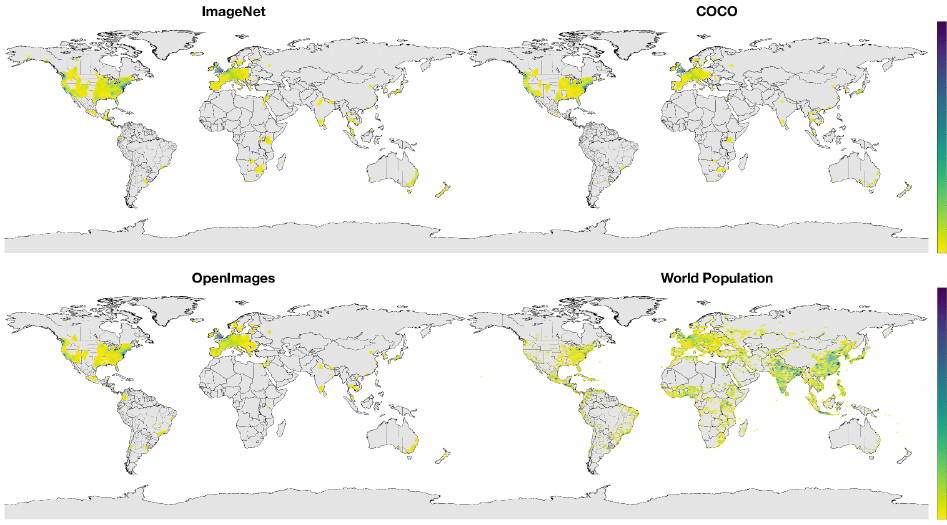}
\caption{Density maps showing the geographical distribution of images in the ImageNet (top-left), COCO (top-right), and OpenImages (bottom-left) datasets. A world population density map is shown for reference (bottom-right).}
\label{fig:density_maps}
\end{figure*}

\item \textbf{Using English as a ``base language'' for data collection.} Most publicly available image-classification datasets were gathered by starting from a list of English words (\emph{e.g.}, all nouns in WordNet~\cite{miller1995}), and performing web searches for images that are tagged with these words on, for example, Flickr. This uni-lingual approach introduces issues because it does not include images that are likely to be tagged in other languages to be included. The aforementioned geographical distribution is one such issue, but more subtle issues can arise. 

For example, certain classes may simply not have an English word associated with them, for instance, particular clothing styles, cultural events, or household items. Alternatively, some languages may be more fine-grained than English in terms of how they define classes (or vice versa). For example, Inuit languages have over a dozen words for ``snow'' to distinguish between different types of snow~\cite{martin1986eskimo,snowwords}. Even if a word exists and means exactly the same thing in English and in some other language, the visual appearance of images associated with that word may be very different between English and the other language; for instance, an Indian ``wedding'' looks very different than an American wedding and Indonesian ``spices'' are very different from English spices. To demonstrate these effects, we performed a series of image searches on Flickr using English nouns and their translations in Hindi. We show representative samples for some of these searches in Figure~\ref{fig:flickr_queries}, and suggest there are clear visual differences between search results for the same query in two different languages.

\end{enumerate}

\section{Related Work}
\label{sec:related_work}
This work is related to a larger body of work on fairness and on building representative computer-vision systems.

\noindent\textbf{Fairness in machine learning.} A range of recent papers have studied how to develop machine-learning systems that behave according to some definition of fairness. Several formulations of fairness exist. For instance, \emph{statistical parity} \cite{calders2009,calders2012,edwards2016,johndrow2017,kamiran2009,kamishima2011,louizos2016,zemel2013} states that in binary-classification settings, members of different groups should have the same chance of receiving a positive prediction. Because statistical parity may be inappropriate when base rates differ between groups, \emph{disparate impact}~\cite{feldman2015,zafar2015} poses that positive-classification rates between any two groups should not vary by more than $80\%$. The \emph{equalized odds}~\cite{hardt2016} fairness principle (also referred to as \emph{disparate mistreatment}~\cite{zafar2017}) requires classification algorithms to make predictions such that no group receives a disproportionately higher number of false-positive or false-negative errors. The \emph{demographic parity} formulation of fairness does not focus on group membership, but are based on the idea that similar individuals should receive similar predictions~\cite{dwork2012}. Selecting and maintaining the ``correct'' fairness requirement for a real-world is no easy task \cite{beutel2019}, in particular, in situations in which the group membership of the system's users is unknown. Moreover, several impossibility results exist: for instance, equalized odds is incompatible with other formulations of fairness~\cite{chouldechova2017,corbett2017,kleinberg2017}, and it is impossible to achieve equalized odds using a calibrated classifier without withholding a randomly selected subset of the classifier's predictions~\cite{pleiss2017}.

The empirical study presented here does not neatly fit into many of the existing fairness as it focuses on multi-class (and potentially multi-label) prediction rather than binary prediction. Moreover, the input provided to image-recognition systems does not contain information on the user or its group membership, which makes it difficult to apply fairness formulations based on group membership of similarities between individuals. Having said that, commonly used techniques to increase fairness, such as instance re-weighting~\cite{jiang2019}, analyzing features~\cite{adler2018auditing} may help in training image-recognition systems that work for everyone.

\noindent\textbf{Building representative computer-vision systems.} Several recent papers have identified and analyzed biases in other types of computer-vision systems. For instance, commercial gender classification systems were found to have substantially higher error rates for darker-skinned females than for light-skinned males~\cite{boulamwini2018,raji2019actionable,phillips1997feret}. A study of Google Image Search revealed exaggeration of stereotypes and systematic underrepresentation of women in search results~\cite{kay2015}, and a study of ImageNet revealed correlations between classes and race~\cite{stock2017}. Other studies have revealed biases in computer-vision datasets that allow models to recognize from which dataset an image originated~\cite{tommasi2017,torralba2011}. Most related to our study is a prior analysis suggesting that for certain classes, the confidence of image classifiers may vary depending on where the image was collected~\cite{shankar2017}.

\section{Discussion}
\label{sec:discussion}
The analysis presented in this paper highlights biases in modern object-recognition systems, but it hardly tells the full story. In particular, our study only addresses two of the five main sources of bias in machine learning systems~\cite{suresh2018}: it addresses \emph{representation bias} and elements of \emph{measurement bias}. It does not address historical bias in the data, or evaluation and aggregation biases that may have implicitly influenced the development of our models.

More importantly, our study has identified geographical and income-related accuracy disparities but it has not solved them. Our analysis in Section~\ref{sec:analyzing_errors} does suggest some approaches that may help mitigate these accuracy disparities such as geography-based resampling of image datasets and multi-lingual training of image-recognition models, for instance, via multi-lingual word embeddings~\cite{conneau2017}. Such approaches may, however, still prove to be insufficient to solve the problem entirely: ultimately, the development of object-recognition models that work for everyone will likely require the development of training algorithms that can learn new visual classes from few examples and that are less susceptible to statistical variations in training data. We hope this study will help to foster research in all these directions. Solving the issues outlined in this study will allow the development of aids for the visually impaired, photo album organization software, image-search services, \emph{etc.}, that provide the same value for users around the world, irrespective of their socio-economic status.

{\small
\bibliographystyle{ieee}
\bibliography{refs}
}

\clearpage
\pagebreak

\section*{Supplemental Material}

\begin{table}[h]
\centering
\small
\resizebox{\linewidth}{!}{
    \begin{tabular}{|c|c|c|}
    \toprule
    Alcoholic drinks & Armchairs & Backyards \\
    Bathroom doors & Bathrooms & Bedrooms \\
    Beds & Bikes & Books \\
    Bowls & Car keys & Cars \\
    Ceilings & Chickens & Child rooms \\
    Cleaning equipment & Computers & Cooking pots \\
    Cooking utensils & Cups & Cutlery \\
    Diapers & Dish brushes & Dish racks \\
    Dish washing soaps & Dishwashers & Drying clothes \\
    Earrings & Everyday shoes & Families \\
    Family snapshots & Favorite home decoration & Floors \\
    Freezers & Front door keys & Front doors \\
    Fruit trees & Fruits & Glasses \\
    Goats & Grains & Guest beds \\
    Hair brushes & Hallways & Hands \\
    Homes & Instruments & Jackets \\
    Jewelry & Kids bed & Kitchen sinks \\
    Kitchens & Latest furniture bought & Light sources \\
    Light sources (living room) & Light sources (kitchen) & Light sources (bed room) \\
    Living rooms & Lock on front doors & Make up \\
    Meat & Medication & Menstruation pads \\
    Mosquito protection & Most loved toys & Motorcycles \\
    Music equipment & Necklaces  & Nicest shoes \\
    Ovens & Palms & Papers \\
    Parking lots & Pens & Pet foods \\
    Pets & Phones & Plates \\
    Plates of food & Power outlets & Power switches \\
    Radios & Refrigerators & Roofs \\
    Rugs & Salt & Shampoo \\
    Shaving & Showers & Soaps \\
    Social drinks & Sofas & Spices \\
    Storage rooms & Stoves & Street view \\
    TVs & Tables with food & Teeth \\
    Toilet paper & Toilets & Tools \\
    Tooth paste & Toothbrushes & Toys \\
    Trash & Vegetable plots & Vegetables \\
    Wall clocks & Wall decorations & Walls \\
    Walls inside & Wardrobes & Washing detergent \\
    Waste dumps & Wheel barrows & Wrist watches\\
    \bottomrule
    \end{tabular}
}
\caption{List of all 117 classes ($20,455$ total images) in the Dollar Street dataset that we used in our analysis of object-recognition systems.}
\label{table:class_list}
\end{table}

\subsection*{License Information for Photos in the Paper}
\par \noindent License information for the photos shown in Figure~\ref{fig:teaser}:
{\small{
\begin{itemize}[leftmargin=*, noitemsep]
\item \underline{''Soap'' UK - Photo}: Chris Dade; Dollar Street (CC BY 4.0).
\item \underline{``Soap'' Nepal - Photo}: Luc Forsyth; Dollar Street (CC BY 4.0).
\item \underline{``Spices'' Philippines - Photo}: Victrixia Montes; Dollar Street (CC BY 4.0).
\item \underline{``Spices'' United States - Photo}: Sarah Diamond; Dollar Street (CC BY 4.0).
\end{itemize}
}}

\par \noindent The photos shown in Figure~\ref{fig:flickr_queries} (in order from left to right, top to bottom) are from Flickr, and have the following licenses:

{
\small{
\begin{itemize}[leftmargin=*, noitemsep]
\item \underline{``Wedding''}: Photo by Elliot Harmon (CC BY-SA 2.0 SA).
\item \underline{``Wedding''}: Photo by Ed Bierman (CC BY 2.0).
\item \underline{``Wedding''}: Photo by Cameron Nordholm (CC BY 2.0).
\item \underline{``Wedding''}: Photo by Victoria Goldveber (CC BY-SA 2.0).
\item \underline{``Wedding in Hindi''}: Photo by Arian Zwegers (CC BY 2.0).
\item \underline{``Wedding in Hindi''}: Photo by Rabia (CC BY 2.0).
\item \underline{``Wedding in Hindi''}: Photo by Abhishek Shirali (CC BY 2.0).
\item \underline{``Wedding in Hindi''}: Photo by Arman Thanvir (CC BY 2.0).
\item \underline{``Wedding in Hindi''}: Photo by Agence Tophos (CC BY 2.0).
\item \underline{``Spices''}: Photo by Collin Anderson (CC BY 2.0).
\item \underline{``Spices''}: Photo by Andrew Malone (CC BY 2.0).
\item \underline{``Spices''}: Photo by Stefan Pettersson (CC BY-SA 2.0).
\item \underline{``Spices''}: Photo by Mike Mozart (CC BY 2.0).
\item \underline{``Spices in Hindi''}: Photo by University of Michigan School for Environment and Sustainability (CC BY 2.0).
\item \underline{``Spices in Hindi''}: Photo by John Haslam (CC BY 2.0).
\item \underline{``Spices in Hindi''}: Photo by Honza Soukup (CC BY 2.0).
\item \underline{``Spices in Hindi''}: Photo by Edward Morgan (CC BY-SA 2.0).
\end{itemize}
}
}
\subsection*{Classes with Largest Accuracy Discrepancy}
\begin{figure}[h]
\centering
\includegraphics[width=\linewidth]{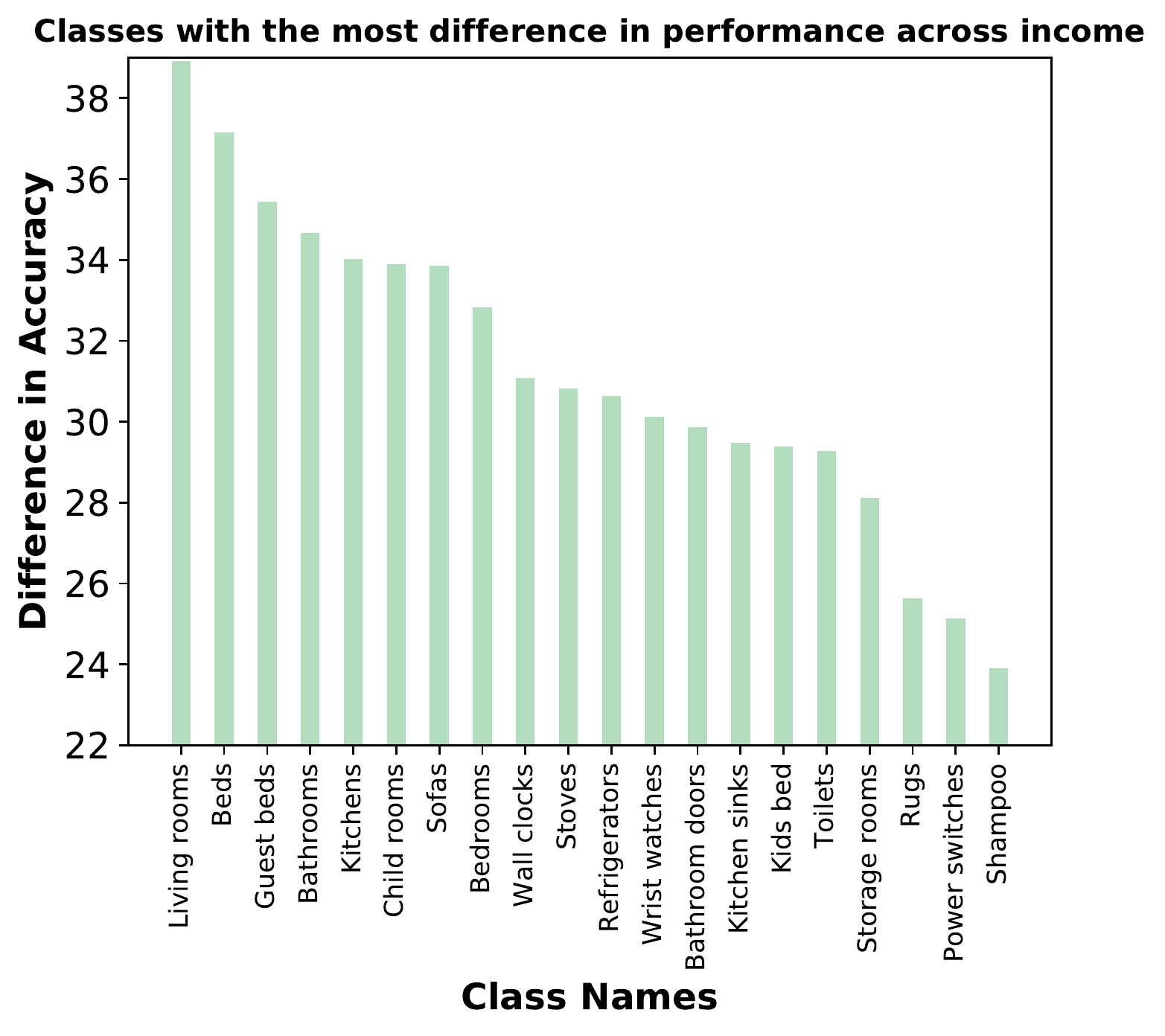}
\vspace{-0.1in}
\caption{Classes for which the difference in accuracy is largest between the highest income bracket and the lowest income bracket.}
\label{fig:diff_accs}
\end{figure}

In Figure~\ref{fig:diff_accs}, we show the 10 classes with the largest discrepancy in average accuracy between the highest income group and the lowest income group. We note that for certain classes, the Dollar Street dataset tends to label images based on their affordance rather than the object. For example, ``refrigerator'' images contain objects such as pots and pans used to cool things for the lower income group images.

\subsection*{Decoupling Geographical Location and Income}
\begin{figure}[h]
\centering
\includegraphics[width=\linewidth]{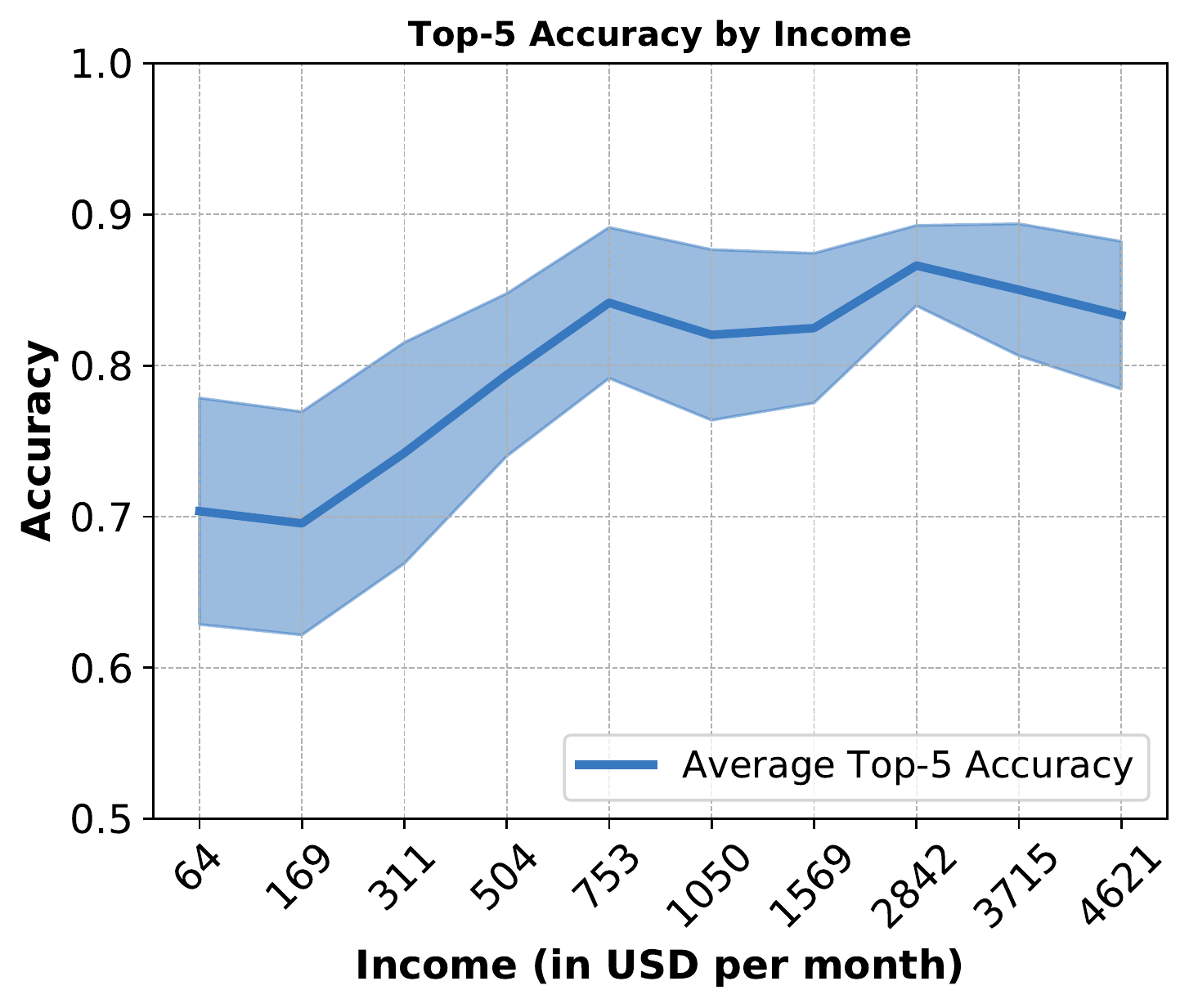}
\caption{Average accuracy of six object-classification systems as a function of the normalized consumption income of the household in which the image was collected (in US\$ per month), measured on all Dollar Street photos taken in India.}
\label{fig:income_india}
\end{figure}

Figure~\ref{fig:income_plots_raw} measured object-recognition accuracy as a function of income on the Dollar Street dataset, which spans a large number of countries. Because geographical location and income are correlated, this raises the question whether location is the sole driver for the observed differences in recognition accuracy. To isolate the effect of income on the accuracy of the recognition systems, we repeat the analysis on Dollar Street images from India -- the country for which most images are available. Figure~\ref{fig:income_india} shows the average accuracy of the six object-recognition systems on the $2,221$ Dollar Street images from India. The income bins in the figure comprise approximately the same number of images ($200 \pm 50$ per bin) as before. Figure~\ref{fig:income_india} reveals a correlation between income and recognition performance, suggesting both location and income partly drive our observations.

\subsection*{Analysis of Facebook System}
\begin{figure}[h]
\centering
\includegraphics[width=\linewidth]{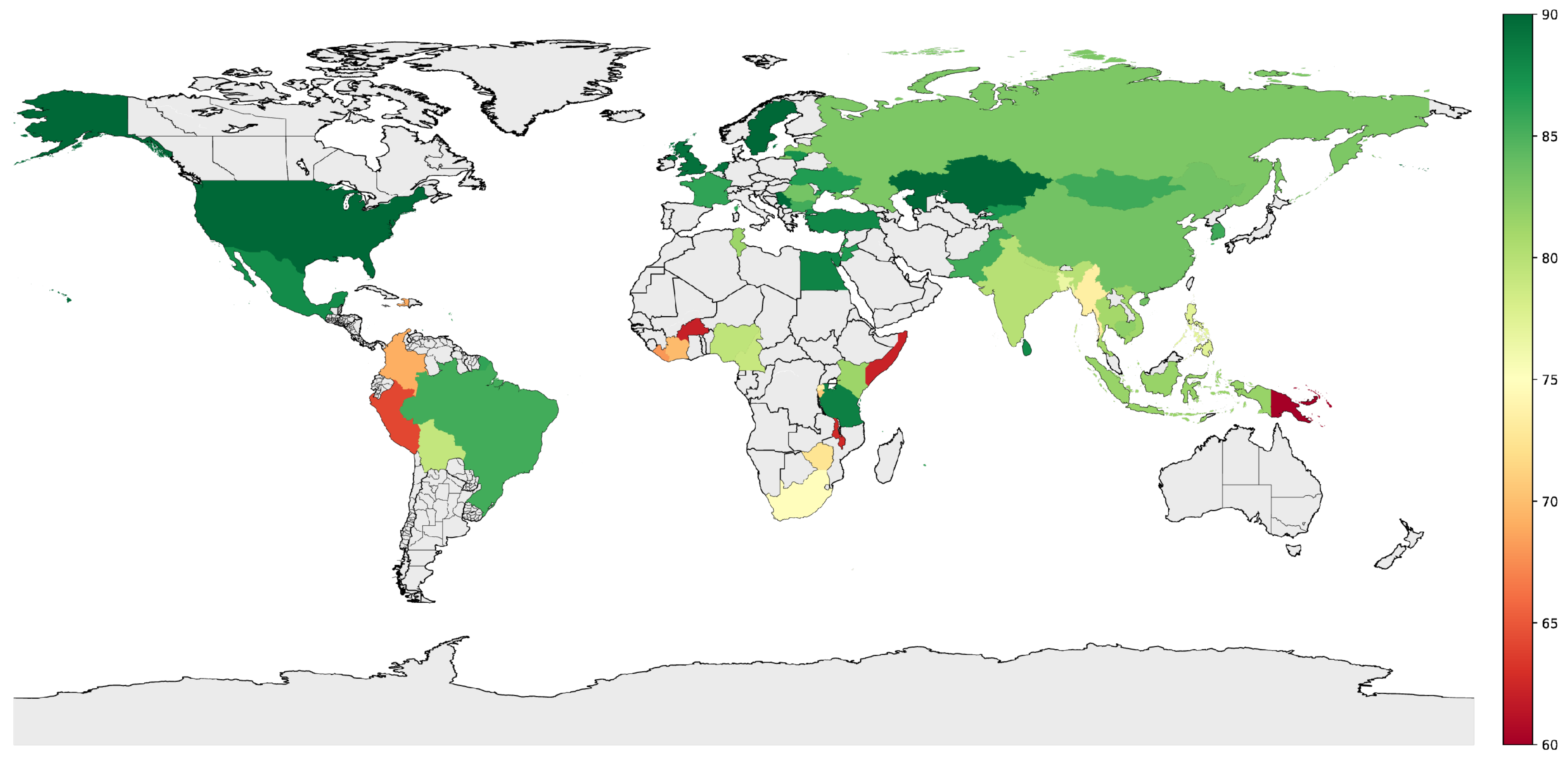}
\caption{Choropleth map displaying the average accuracy of a proprietary Facebook object-recognition systems per country. The color red indicates an accuracy of $\sim\!60\%$, yellow an accuracy of $\sim\!75\%$, and green an accuracy of $\sim\!90\%$.}
\label{fig:facebook_chloropleth}
\end{figure}

Figure~\ref{fig:facebook_chloropleth} shows the accuracy per country of a proprietary Facebook system for object recognition. The results are comparable to those presented in Figure~\ref{fig:rekog_chloropleth}.

\end{document}